Federico Ghelli Visi | mail@federicovisi.com | School of Music in Piteå, Luleå University of Technology, Sweden
Atau Tanaka | a.tanaka@gold.ac.uk | Goldsmiths, University of London, UK

# Interactive Machine Learning of Musical Gesture

Federico Ghelli Visi and Atau Tanaka

## 1 Introduction

This chapter presents an overview of Interactive Machine Learning (IML) techniques applied to the analysis and design of musical gestures. We go through the main challenges and needs related to capturing, analysing, and applying IML techniques to human bodily gestures with the purpose of performing with sound synthesis systems. We discuss how different algorithms may be used to accomplish different tasks, including interacting with complex synthesis techniques and exploring interaction possibilities by means of Reinforcement Learning (RL) in an interaction paradigm we developed called Assisted Interactive Machine Learning (AIML). We conclude the chapter with a description of how some of these techniques were employed by the authors for the development of four musical pieces, thus outlining the implications that IML have for musical practice.

Embodied engagement with music is a key element of musical experience, and the gestural properties of musical sound have been studied from multiple disciplinary perspectives, including Human-Computer Interaction (HCI), musicology, and the cognitive sciences [1]. Likewise, designing gestural interactions with sound synthesis for musical expression is a complex task informed by many fields of research. The results of laboratory studies of music-related body motion based on sound-tracing were indicated as a useful starting points for designing gestural interactions with sound [2]. Informed by environmental psychology, the notion of sonic affordance was introduced to look at how sound invites action, and how this could potentially aid the design of gestural interfaces [3].

Designing and exploring gestural interactions with sound and digital media is at the foundation of established artistic practices where the performer's body is deeply engaged in forms of corporeal interplay with the music by means of motion and physiological sensing [4]. Gesture and embodiment become the core concepts of extended multimedia practices, where composition and interaction design develop side by side [5, 6], and gesture is a fundamental expressive element [7].

### 1.1 Why Machine Learning Musical Gestures? Needs and Challenges

Designing gestural interactions that afford dynamic, consistent, and expressive articulations of musical sound is a challenging and multifaceted task. A key step of the design process is the definition of mapping functions between gesture tracking signals (usually obtained through some motion sensing device) and sound synthesis parameters [8]. These parameter spaces can be very complex,

Federico Ghelli Visi | mail@federicovisi.com | School of Music in Piteå, Luleå University of Technology, Sweden
Atau Tanaka | a.tanaka@gold.ac.uk | Goldsmiths, University of London, UK

depending on the motion sensing and sound synthesis approaches adopted. An effective mapping strategy is one of the crucial factors affecting the expressive potential of a gestural interaction, and as the spaces defined by motion signals and synthesis parameters become more highly-dimensional and heterogeneous, designing mappings can be an increasingly elaborate task, with many possible solutions [9].

In this scenario, gestural interaction design is a robust nontrivial problem, and Machine Learning (ML) techniques can be used by researchers and artists to tackle its complexity in several ways. One of the most notable implications of using ML in this domain is that mappings between gesture and sound can be interactively "shown" to a system capable of "learning" them [10] instead of being manually coded, which in certain situations could become excessively complex and time consuming. In other words, this delineates an interaction design paradigm where interactive systems shift from executing rules to learning rules from given examples [11]. This has advantages in collaborative and interdisciplinary creative practices, as it makes trying and workshopping different gestural interactions easier and quicker, and enables practitioners that are unfamiliar with programming to prototype their own gestural interactions. Moreover, software tools such as the Wekinator [12] and ML libraries for popular programming environments in the arts [13] have made ML for gestural interaction more accessible and easier to learn. Another advantage of interaction design approaches based on ML is that these are often more resilient to noisy and complex input signals than manually programmed mappings. This is particularly useful with certain motion tracking technologies and physiological sensors (see section 2). Noise is not, however, the only challenge when tracking and analysing body movement for musical interaction. Motion tracking systems may return considerably different data when the user changes, different motion-sensing technologies measure and represent movement in very different ways, musical gestures may convey musical ideas at different timescales [14] and therefore it should be possible to model both spatial and temporal features of musical gestures while maintaining the possibility of dynamic and continuous expressive variations. We will now describe how ML is an helpful resource in addressing these challenges.

### 1.2   Chapter Overview

The sections that follow will describe the main components of an IML system for gesture-sound interaction – schematised in Figure 1 – namely motion sensing, analysis and feature extraction, ML techniques, and sound synthesis approaches. Following this, we will describe the typical workflow for deploying an IML system for gesture-sound mapping and how this model can be extended further using RL to explore mapping complexity in an AIML system prototype. We will then describe how this models were used in some pieces composed by the authors, before closing the chapter with some remarks regarding the necessity of adopting an interdisciplinary approach encompassing basic research, tools development, and artistic practice in order to make substantial advances in the field of expressive movement interaction. We finish by showing that the research field has

Federico Ghelli Visi | mail@federicovisi.com | School of Music in Piteå, Luleå University of Technology, Sweden
Atau Tanaka | a.tanaka@gold.ac.uk | Goldsmiths, University of London, UK

implications stretching way beyond the music domain, given the increasing role of ML technologies in everyday life and the peculiarities that make music and the arts a fertile ground for demystifying ML and thereby understanding ways of claiming and negotiating human agency with data and algorithmic systems.

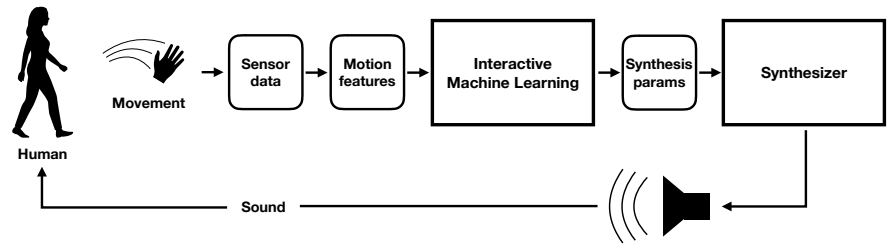

**Fig. 1.** Architecture of an Interactive Machine Learning system for gestural musical interaction.

## 2 Machine-sensing Gesture

Capturing body movement for the purpose of real-time interaction with sound may be done by various technological means. Rather than providing a list of the many devices available for this purpose, we will describe the main approaches for tracking body movement employed in the context of music and multimedia performance, and the implications that adopting an approach over another has for using ML techniques. These include notes on how different types of motion data represent movement, and the opportunities afforded by the use of physiological data.

### 2.1 Sensing Movement

**Optical Sensing**

Optical motion sensing relies on the analysis of the signals coming from various kinds of video cameras. There are many examples of multi-camera systems used in the arts [15, 16] as well as of systems using more sophisticated optical approaches such as depth and stereoscopic cameras.

Despite the technology being a few decades old, marker-based infra-red Motion Capture (MoCap) is still considered as one of the most reliable methods for measuring complex movement in a three-dimensional space. Tracking precision and temporal resolution have progressively improved, allowing accurate tracking of finger movements and facial expressions. Recent MoCap systems are also capable of streaming motion data live, thus making real-time applications possible. Data obtained from these systems is usually in the form of three-dimensional vectors referring to a global coordinate system. Each sample in the data returns three-dimensional information regarding the position of a point (marker) in space in relation to the origin of the Cartesian axes. The origin is defined during the calibration procedure and is usually set in an arbitrary place on the floor within the capture area. Most marker-based systems also allow to track movement in six degrees of freedom (6DoF), meaning that – in addition to position along the three spatial axes – the system also returns information on the orientation and rotation of a point in space along three rotational axes. This information is usually represented in Euler angles or quaternions. In MoCap systems, 6DoF tracking is usually achieved by processing positional data of single markers grouped into a rigid body in a predefined spatial configuration. This should be unique for each rigid body in order to avoid mislabelling when multiple rigid bodies are in the capture space at the same time.

**Inertial Measurement Units**

Inertial Measurement Units (IMU) are small, low-cost, highly portable devices that incorporate accelerometers and gyroscopes. When these devices are paired with magnetometers, the resulting arrays are also known as Magnetic, Angular Rate and Gravity (MARG) sensors. These sensor arrays allow the tracking of acceleration, rotational velocity and orientation relative to the earth's magnetic field of whatever they are attached to. They are used extensively in aviation, robotics and HCI. Their increasing affordability and small size have made them a very common feature of mobile and wearable devices and other consumer electronics. Sensors featuring 3D accelerometers, 3D gyroscopes, and 3D magnetometers have become the most widely used type of IMU/MARG. They enable to estimate various motion features including optimised three-dimensional orientation obtained by fusing together the data from the different types of sensors. These devices are often marketed as 9DoF (9 Degrees of Freedom) sensors, since they consist of three tri-axis sensors and thus have a total of nine sensitive axes.

Whereas the raw data obtained using marker-based optical motion capture consists of samples of position based on a 3D Cartesian coordinate system, the data returned by IMU/MARG sensors is usually in the form of three three-dimensional vectors, each one expressing acceleration, rotational velocity, and orientation respectively. Calculating absolute position from the data of a single IMU in real time is technically very difficult if not unfeasible, as the operation would require double integration of acceleration data. This would result in a considerable amount of residual error since drift would accumulate quadratically.

The lack of reliable information on absolute position when using single IMUs is a key difference between data obtained through inertial sensing and that of optical motion capture. The data obtained from IMUs sensors is morphologically very different from positional data returned by optical MoCap. The differences in the way movement is tracked and represented by the two different technologies have implications on how movement data is eventually interpreted and used,

Federico Ghelli Visi | mail@federicovisi.com | School of Music in Piteå, Luleå University of Technology, Sweden
Atau Tanaka | a.tanaka@gold.ac.uk | Goldsmiths, University of London, UK

particularly in the context of expressive movement tracking and ML of musical gestures. As an example, single IMUs afford working with movement relative to the body of the performer and postures, whereas having access to absolute positions may enable interaction strategies that take into considerations the spatial relationships between different performers and the different areas of the performance space where the action is taking place.

### 2.2 Sensing the Body

It can be argued that representing human movement solely as displacement of body parts in a three-dimensional space would result in a limited interpretation. Merleau-Ponty maintains that we act upon the environment through proprioception and "a knowledge bred of familiarity which does not give us a position in objective space" [17, p. 166]. Salazar Sutil [18] points out that the conceptualisation of corporeal movement is often optically biased, whereas sensations that are independent from sight are often neglected. Thus, we argue that expressive body movement cannot be entirely represented and therefore fully understood exclusively by means of visual media. In the context of music performance, we looked at the concepts of intention, effort, and restraint in relation to the use of electromyogram (EMG) for digital musical instrument application [19]. EMG is a signal representing muscle activity employed in the biomedical and HCI fields as a highly sensitive way to capture human movement and has been used as a signal with which to sense musical gesture [20, 21]. Using EMG for music presents several challenges. The raw signal itself resembles noise and sensing such a low voltage signal is difficult to do without accumulating noise from the environment. Individual anatomies vary and we each employ our muscles differently, even when performing what looks like the same gesture. Basic signal processing can only go so far when interpreting expressive, nuanced biosignals. Adopting approaches based on ML can considerably help with these challenges, making EMG an attractive technology for musical interaction. In particular, supervised learning approaches – which will be described in section 4 – constitute a way for tackling the intersubjective variability and the noisy quality of muscular signals.

## 3 Analysing Gesture

Higher-level descriptors are often used to extract features from raw motion data to help describing body movement in a more meaningful way. Such descriptors are frequently employed in expressive movement analysis, motion recognition, and music performance analysis. Feature extraction is a crucial step in an IML pipeline. This is an important task, as it will affect how ML algorithms will interpret body movement and therefore determine the affordances of the resulting gesture-sound interactions.

Programming environments such as Eyesweb [22] offer solutions dedicated to real-time human movement analysis and feature extraction. Libraries for real-time motion analysis such as the Musical Gesture Toolbox were initially

Federico Ghelli Visi | mail@federicovisi.com | School of Music in Piteå, Luleå University of Technology, Sweden
Atau Tanaka | a.tanaka@gold.ac.uk | Goldsmiths, University of London, UK

dedicated mainly to standard RGB video analysis [23]. The library has been developed further to process MoCap data and be compatible with several programming environments [24]. Notably, some of the features that were initially designed for analysing video data – such as Quantity of Motion (QoM, see section 3.1) – have been extended for the use with MoCap data. We developed the *modosc* library to make methods for handling complex motion data streams and compute descriptors in real time available in music performance systems [25, 26]. At the time of writing, the library is being extended for the use with IMU and EMG in addition to MoCap data. The following sections will give an overview of some of the descriptors most widely used for processing motion and EMG data.

### 3.1 Motion Features

**Fluidity**

Inspired by the theoretical work on human motion by Flash and Hogan – which maintains that trajectories of human limbs can be modelled by the *minimum jerk law* [27], Piana et al. [28] defined Fluidity Index as the inverse of the integral of jerk. Jerk - or "Jolt" - is the third-order derivative of position, i.e. the rate of change of the acceleration of an object with respect to time. Fluidity Index has been used with supervised learning algorithms for the purpose of recognising expressed emotions from full-body movement data [28].

**Quantity of Motion**

Fenza et al. defined Quantity of Motion (QoM) as the sum of the speeds of a set of points multiplied by their mass [29]. Glowinski et al. [30] included a similar measure in their feature set set for the representation of affective gestures, denoted as "overall motion energy." This motion feature has also been used for real-time video analysis [23] and a version for IMU data was also proposed [5].

**Contraction Index**

Contraction Index is calculated by summing the Euclidean distances of each point in a group from the group's centroid [29]. It is an indicator of the overall contraction or expansion of a group of points and – similarly to Fluidity Index – it has been used for emotion recognition applications [28].

When using independent inertial sensors, the lack of positional data might make it difficult to compute Contraction Index. An alternative measure of contraction and expansion of body posture using IMU data was proposed by Visi et al. [5]. This solution uses the Euclidean distance between projected points to estimate whether the limbs of a person wearing IMUs are pointing in opposite directions.

**Bounding Shapes**

Bounding shapes have been used in the analysis of affective gestures [30] as well as in dance movement asnalysis [31]. Several bounding shapes can be used

Federico Ghelli Visi | mail@federicovisi.com | School of Music in Piteå, Luleå University of Technology, Sweden
Atau Tanaka | a.tanaka@gold.ac.uk | Goldsmiths, University of London, UK

for real-time movement analysis. For example, a *bounding box* is the rectangular parallelepiped enclosing a given group of points in a 3d space. Assuming these points are placed on the body of a performer, the height, width, and depth of the bounding box can be used as an indicator of the posture of the full body evolves over time. The minimum polyhedron that encloses a given group of points in a 3D space is instead called *three-dimensional convex hull*. The volume of the convex hull represents the size of the space the body interacts with, and can be used as a feature for various ML tasks.

**Periodic Quantity of Motion**

Periodic Quantity of Motion (PQoM) was proposed as a way to measure periodicity in the movement in relation to the musical rhythm [5], or – in other words – how much body movement resonates with each rhythmic subdivision (i.e. quarter note, eighth note, etc.). The first PQoM implementation was designed to extract periodic motion from optical motion capture data [32]. The PQoM is estimated by decomposing the motion capture signal into frequency components by using filter banks [33]. The amplitude of the signal for each frequency component corresponds to an estimate of the resonance between the corresponding rhythmic subdivision and the movement. A script for PQoM estimation was made available as an extension to version 1.5 of the MoCap Toolbox for Matlab [34], and a newer, redesigned version of the script has recently been made available [35].

### 3.2 EMG Features

**Signal Amplitude**

One of the most important features of EMG signals is the amplitude of the signal with respect to time. This measure is related to the force exerted while executing a gesture. Given the complexity and variability of the EMG signal, reliable amplitude estimation may be challenging. Simply applying a low-pass filter to the signal to reduce undesired noise may result in the loss of sharp onsets describing rapid movement and may also introduce latency when processing the signal in real time. Adopting a nonlinear recursive filter based on Bayesian estimation [36] significantly reduces the noise while allowing very rapid changes in the signal, greatly improving the quality of the signal for real-time gestural interaction.

**Mean Absolute Value**

Mean Absolute Value (MAV) is one of the most popular features used in EMG signal analysis [37]. It has been shown that MAV is more useful than other features for gesture recognition tasks based on supervised learning algorithms [38]. MAV corresponds to the average of the absolute values of the EMG signal amplitudes in a given time window. When computed in real-time, a larger time window returns a smoother signal, whilst a shorter one can be useful to track sharper onsets in muscular activity.

Federico Ghelli Visi | mail@federicovisi.com | School of Music in Piteå, Luleå University of Technology, Sweden
Atau Tanaka | a.tanaka@gold.ac.uk | Goldsmiths, University of London, UK

8          Visi & Tanaka

**Root Mean Square**

Root mean square (RMS) is a popular signal processing feature, widely used for audio analysis. With EMG signals, it has been used together with ML algorithms for gesture classification tasks [39]. RMS is equals to the square root of the sum of the squares of the values of the signal in a given time window.

**Teager-Kaiser Energy-tracking Operator**

The Teager-Kaiser Energy-tracking Operator (TKEO) was first proposed as a way for estimating energy in speech signals [40]. It has been employed for a variety of signal processing tasks, including noise suppression [41]. It has been shown that TKEO considerably improves the performance of onset detection algorithms also in situations with a high signal-to-noise ratio [42]. The feature can be be easily calculated from three adjacent samples. For each signal sample, TKEO is equal to the square of the amplitude minus the product of the precedent and successive samples.

**Zero Crossing Rate**

The Zero Crossing Rate (ZCR) corresponds to the number of times the signal changes sign within a given time window. Another widely used feature in audio signal processing, it is used to recognise periodic sounds from noisy ones, and it is employed in speech recognition [43]. Caramiaux et al. [44] use ZCR as one of the features for the analysis of two different modalities of muscle sensing to explore the notion of gesture power.

## 4  Machine Learning Techniques

ML techniques are statistical analysis methods and computational algorithms that can be used to achieve various tasks by building analytical models (i.e. "learning") from example data. Many ML technique involve a *training* phase and a *testing* phase. During the training phase, sample data is used to model how the system should respond and perform different tasks. During the *testing* phase, new input data is fed into the model, which then responds and performs tasks following decisions based on structures and relationships learnt during the training phase. As an example, during the training phase a performer using motion sensors may want to record a gesture and associate it to specific sounds being produced by a sound synthesis engine. Then during the testing phase, the performer moves freely while the system follow their movements and infer which sounds should be played according to the examples given during the training phase. This allows for flexibility and generalisation, making ML techniques particularly useful for complex applications that involve many variables and that may be dependent on factors that are difficult to predict or control, such as the environments in which systems are deployed, or high variability in how the system responds to different users. For example, in a musical context one may want to use a gesture-sound interaction system in different performance spaces,

Federico Ghelli Visi | mail@federicovisi.com | School of Music in Piteå, Luleå University of Technology, Sweden
Atau Tanaka | a.tanaka@gold.ac.uk | Goldsmiths, University of London, UK

which may have different lighting conditions. This may result in undesirable unexpected behaviours, such as the system responding differently in the concert hall where a piece is to be performed compared to the space where the piece has been rehearsed. Moreover, the system may be used by different performers, whose bodies may differ considerably and thus be tracked differently by various types of sensors (see section 2). In such situations designing interactions by explicitly programming how each sound parameter should behave in response to incoming sensor data might be too time-consuming, impractical, or result in interactions that are too shallow and do not afford expressive variations.

There are several standard learning strategies to train a program to execute specific tasks [45]. Among the most common paradigms, we find Supervised Learning (SL), Unsupervised Learning (UL), and Reinforcement Learning (RL). In SL, the training data consists of input paired with desired output. In other words, training examples are *labelled*. For example, in a supervised learning scenario motion feature data is paired with the desired sound and passed to the learning algorithm as training data. Classification and regression are some of the most common supervised learning tasks In UL, training data is unlabelled. The goal is learnt from the data itself, by analysing patterns and underlying structures in the given examples. As an example, a set of unlabelled sounds may constitute the training set and the task of the unsupervised learning algorithm may be to group the sounds that have similar features. Common unsupervised learning tasks include clustering, and dimensionality reduction. In a study by Visi et al. [46] dimensionality reductions approaches are employed to observe commonalities and individualities in the music-related movements of different people miming instrumental performances. The peculiarity of strategies based on RL is that the algorithm is given feedback in response to the actions this has executed. The goal of the algorithm is to maximise the positive feedback – or rewards – they are given by a human (or by another algorithm) that is observing the outcome of their actions. Training and testing phases here are more intertwined than in typical supervised and unsupervised strategies, as training occurs through testing. For example, in a RL scenario, one may task an algorithm to propose some sound synthesis presets, and the user may give positive or negative feedback in order to obtain a sound that is closer to their liking. Parameter space exploration is a task associated with this learning strategy. A gesture-sound mapping exploration method that takes advantage of RL [47] will be described in section 6.

The following sections will outline how these strategies are employed to perform tasks often associated with ML of musical gesture, namely classification, regression, and temporal modelling.

### 4.1 Classification

Classification is the task of assigning a category, or class, to an item. In a supervised learning scenario, the training dataset is constituted by items labelled with the category they belong to. The training dataset is then used to build a model that will assign labels to new unlabelled items, or instances, that have not been classified before. As an example in the context of musical gestures, the training set may be made of discrete gestures (e.g. tracing a circle in the air, or a triangle...) where the sensor data and motion features resulting from performing such gestures are paired with corresponding label (circle, triangle, etc.). These labelled gestures constitute a *vocabulary*. In performance, the classifier may be used to track the movements of the performer and recognise when one of the gesture in the vocabulary is being performed. Successful recognition of one of the gestures in the vocabulary may be then paired with specific musical events (e.g. play a kick drum sample when the tracked gesture is classified as a circle, play a snare sample when the gesture is classified as a triangle, etc.). In a typical gesture classification scenario, classification occurs *after* the gesture is performed, and output of the model is *discrete*, meaning that a gesture will always belong to one of the defined classes. Common algorithms for classification include K-Nearest Neighbors (k-NN), Adaptive Boosting (AdaBoost), Support Vector Machines (SVM), and Naive Bayes. These and other algorithms are described in detail in the manual by Hastie et al. [48]. It is important to note that different classification algorithms afford different interaction sound parameter mapping approaches. For example, by using a probabilistic classifier such as Naive Bayes, one can use the probability distribution (i.e. the set of likelihoods that the incoming gesture belongs to each of the predefined classes) and map their values to parameters (e.g. a set of volume levels) instead of using the class labels to trigger discrete musical events. Finally, classifiers can be used to recognise static postures if trained – for example – with data describing absolute or relative position of parts of the body of a performer. Classification of gestures based on how they unfold over time can be done by using various temporal modelling approaches, which will be described in section 4.3.

### 4.2 Regression

Regression is the task of estimating the relationship between an independent variable (or a *feature*) and a dependent, or *outcome*, variable. This is done by building a statistical model that explains how the variables are related, and thus allows to infer the value of the dependent variable given the independent variable. The model describing this continuous function is built using a set of discrete samples of independent variables (the input) paired with the corresponding values of the dependent variables (the output). Building a regression model is a supervised learning problem, given that to do so one requires labelled data (input paired with corresponding output). Regression is used in several domains for tasks such as prediction and forecasting. In the context of musical interaction, regression is an attractive approach as it allows to define complex, continuous mapping functions between gesture features and sound synthesis parameters. This can be done by providing examples consisting in sample input data (e.g. motion or EMG features, see section 3) paired with sound synthesis parameter.

Artificial Neural Networks (ANN) are an efficient way to build linear regression models. A typical ANN is a network of binary classifiers – called perceptrons – organised in a number of layers. Perceptrons are also referred to as "neuorns" or

Federico Ghelli Visi | mail@federicovisi.com | School of Music in Piteå, Luleå University of Technology, Sweden
Atau Tanaka | a.tanaka@gold.ac.uk | Goldsmiths, University of London, UK

"nodes." The first layer of the network (the input layer) has a node for each input feature. Perceptrons in layers after the input layers produce an output based on the activation function (the binary classifier, generally a sigmoid function, but other activation functions may be used) applied to the input they received from the previous layer. The function includes a set of weights applied to each input and an additional parameter, the bias. After producing the output of each node feeding layer after layer (the feed forward process), error is calculated and a correction is sent back in the network in a process known as back propagation. After a number of iterations, or epochs, error is progressively reduced. ANNs are an attractive ML technique when dealing with real-time motion tracking, as they can handle errors in the incoming data (which may be caused by noisy sensor signal) relatively well.

The model obtained by training a neural network may then be used to map incoming motion features to sound synthesis continuously and in real time. Several approaches based on regression may be used to map gestural features to sound synthesis [20]. We have developed the GIMLeT pedagogical toolkit for Max [49] to provide some practical examples of using linear regression for this purpose. However, ordinary ANNs do not take into account temporal aspects of the input data. The next section will look at some of the approaches designed to analyse and follow the evolution of a gesture in time.

### 4.3 Temporal Modelling

Gesture unfolds over time, and gestures that may look similar in terms of displacement in space may differ radically in expressivity depending on their temporal evolution. For example, moving an arm outwards very slowly or very fast following the same trajectory may convey very different expressive intentions. While certain types of neural networks such as Echo State Networks exhibit short-term memory and can be trained to operate on temporal aspects of their input [50], longer time spans require different approaches. Dynamic Time Warping (DTW) [51] is a technique that allows to temporally align incoming time series (e.g. motion features changing over time) to previously saved gesture templates. Templates are pre-recorded gesture examples. The DTW algorithm will attempt to align incoming gesture features to the set of recorded gesture templates, also referred to as a gesture vocabulary. This way, it is possible to perform various tasks including assessing to which gesture template the incoming motion data is closer to. DTW has been used extensively for music applications such as musical gesture classification [52], to evaluate timing of musical conducting gestures [53], or as a distance measure to place musical gestures in a feature space [46]. One major drawback of DTW for musical applications is that, albeit giving access to how a gesture evolves over time, recognition occurs only after the gesture has been fully performed, and thus not continuously. To address this limitation, Bevilacqua et al. [54] proposed a real-time gesture analysis system based on Hidden Markov Models (HMM). This method allows to continuously recognise a gesture against stored gesture templates, outputting parameters describing time progression (i.e. how much of the gesture has already been performed, this is known as "gesture following") and the likelihood of the gesture belonging to one of the predefined gesture classes. This allows musical interactions such as audio stretching/compressing in synchronisation with gesture performance. Françoise et al. [55] extended this approach further, proposing a set of probabilistic approaches to define motion to sound relationships. These include a hierarchical structure that allows to switch between the difference gestures in the vocabulary and follow the temporal progression of the likeliest matching template while performing, and a multimodal approach that models the temporal evolution of both the motion features and the sound parameters. Caramiaux et al. [56] proposed further extensions to continuous gesture following by focusing on the online analysis of meaningful variations between gesture templates and performed gestures. Their approach uses particle filtering for tracking variations from the recorded template in real time, allowing to estimate geometric variations such as scaling (i.e. how much is the gesture bigger/smaller than the template?), rotation angles (i.e. how much is the performed gesture tilted in respect to the template?), and temporal dynamics (i.e. is the gesture performed faster or slower than the recoded template?). This gesture variation parameters can then be mapped to sound synthesis parameters. For example, the authors describe a study where an increase in scaling corresponds to louder volume, temporal dynamics are mapped to playback speed of samples, and rotation angles to high-pass filtering [56, p. 19].

## 5 Sound Synthesis and Gesture Mapping

Modern sound synthesis techniques are often characterised by a high number of parameters one can manipulate in order to make different sounds. Whilst these afford vast synthesis possibilities, exploring the resulting extensive parameter spaces may be a challenging task, which can be particularly difficult to accomplish by manipulating every parameter by hand.

The choice of synthesis algorithm, therefore, can be one where individual synthesis parameters may be difficult to manually parametrise by hand. Instead, we will exploit the "mapping by demonstration" paradigm where the ML algorithm will create a model whereby performance input is translated to synthesis output. In this regard, a difficult to programme synthesis method like Frequency Modulation, could be a good candidate.

Here we present two approaches from our work to demonstrate how regression can work with different levels of complexity of sound synthesis. We show a simple granular synthesiser, and a more sophisticated synthesis using content-based concatenative synthesis [57].

### 5.1 Granular Synthesis and Sound Tracing

We created a basic granular synthesis module using the in-built capabilities of a sample buffer reader in Max [58], `groove~`. The implementation is a time domain sample-based synthesiser where an audio buffer contains the sample being played,

Federico Ghelli Visi | mail@federicovisi.com | School of Music in Piteå, Luleå University of Technology, Sweden
Atau Tanaka | a.tanaka@gold.ac.uk | Goldsmiths, University of London, UK

and pitch transposition and playback speed are decoupled. This is coupled with subtractive synthesis with a classic resonant low-pass filter. There are six control parameters:

- playback start time
- playback duration
- playback speed
- pitch shift
- filter cutoff frequency
- filter resonance.

A version of this synthesiser is available as part of the GIMLeT pedagogical toolkit for Max [49]. This provides an example of a synthesiser where the sound authoring parameters are human readable and where parametrisation could be done by hand. The challenge comes in creating sounds that dynamically respond to incoming gesture without hardwiring gestural features to synthesis parameters in a traditional mapping exercise. Here we use sound tracing [59] as a method where a sound is given as a stimulus to create evoked gestural response. By gesticulating to a sound that evolves in time, we author gesture that then becomes training data for the regression algorithm in a "mapping-by-demonstration" workflow. In order to author time varying sound using this synthesiser, we create a system of "anchor points", salient points in the timbral evolution of the sound that are practical for sound synthesis parametrisation, and useful in pose-based gesture training [20]. The synthesiser is controlled by our break-point envelope-based playback system and enables the user to design sounds that transition between four fixed anchor points (start, two intermediate points, and end) that represent fixed synthesis parameters. The envelope interpolates between these fixed points. The temporal evolution of sound is captured as different states in the breakpoint editor whose envelopes run during playback, feeding both synthesiser and the ML algorithm. Any of the synthesis parameters can be assigned to break-point envelopes to be controlled during playback.

These sound trajectories are then reproduced during the gesture design and model training phases of our workflow. In performance a model maps sensor data to synthesis parameters, allowing users to reproduce the designed sounds or explore sonic space around the existing sounds.

### 5.2 Corpus-based Synthesis and Feature Mapping

Corpus-based concatenative synthesis (CBCS) is a compelling means to create new sonic timbres based on navigating a timbral feature space. In its use of atomic source units that are analysed, we can think of CBCS as an extension of granular synthesis that harnesses the power of music information retrieval and the timbral descriptors they generate. The actual sound to be played is specified by a target and features associated with that target.

A sound file is imported into the synthesiser, and it is automatically segmented into units, determined by an onset segmentation algorithm. A vector of

Federico Ghelli Visi | mail@federicovisi.com | School of Music in Piteå, Luleå University of Technology, Sweden
Atau Tanaka | a.tanaka@gold.ac.uk | Goldsmiths, University of London, UK

19 auditory features, shown in Table 5.2, are analysed for each unit. Playback typically takes place as a navigation in the audio feature space. A set of desired features is given to the synthesiser, and a k-nearest neighbours algorithm retrieves the closest matching unit to a given set of auditory features. This synthesis method, therefore, is not one where the user programmes sound by setting synthesis parameters in a deterministic way. Schwarz typically uses CataRT[60] controlled through a 2D GUI in live performance, which enables control of only two target audio features at a time.

Here, the full vector of all auditory features are associated with with the sensor feature vectors to train the neural network, and roughly represent a high-dimensional timbral similarity space. We refer to this as multidimensional *feature mapping*, that is to say, that a gesture-sound mapping is created in the feature domain.

**Table 1.** Audio feature vector for corpus based concatenative synthesis

| Sound Features |
| --- |
| Duration |
| Frequency $\mu$ |
| Frequency $\sigma$ |
| Energy $\mu$ |
| Energy $\sigma$ |
| Periodicity $\mu$ |
| Periodicity $\sigma$ |
| AC1 $\mu$ |
| AC1 $\sigma$ |
| Loudness $\mu$ |
| Loudness $\sigma$ |
| Centroid $\mu$ |
| Centroid $\sigma$ |
| Spread $\mu$ |
| Spread $\sigma$ |
| Skewness $\mu$ |
| Skewness $\sigma$ |
| Kurtosis $\mu$ |
| Kurtosis $\sigma$ |

The high- dimensionality of gesture and sound feature spaces raises challenges that ML techniques have helped to tackle. However, this complexity also offers opportunities for experimentation. This led us to develop an extension to the IML paradigm that allows to explore the vast space of possible gesture-sound mappings with the help of an artificial agent and RL.

Federico Ghelli Visi | mail@federicovisi.com | School of Music in Piteå, Luleå University of Technology, Sweden
Atau Tanaka | a.tanaka@gold.ac.uk | Goldsmiths, University of London, UK

## 6 Reinforcement Learning

RL is an area of ML in which algorithms in the form of artificial agents are programmed to take actions in an environment defined by a set of parameters. Their goal is to maximise the positive feedback – or rewards – they are given by a human (or by another algorithm) observing the outcome of their actions. Deep RL approaches – such as the Deep TAMER algorithm – leverage the power of deep neural networks and human-provided feedback to train agents able to perform complex tasks [61]. Scurto et al. [62] implemented the Deep TAMER algorithm to design artificial agents that allow to interactively explore the parameter spaces of software synthesisers.

We have developed a system that makes use of Deep RL to explore different mappings between motion tracking and a sound synthesis engine [47]. The user can give positive or negative feedback to the agent about the proposed mapping while playing with a gestural interface, and try new mappings on the fly. The design approach adopted is inspired by the ideas established by the IML paradigm (which we schematised in Figure 1), as well as by the use of artificial agents in computer music for exploring complex parameter spaces [63–65]. We call this interaction design approach *Assisted Interactive Machine Learning* (AIML).

### 6.1 RL for Exploring Gesture-sound Mappings: Assisted Interactive Machine Learning

An AIML system is designed to interactively explore the motion-sound mappings proposed by an artificial agent following the feedback given by the performer. This iterative collaboration can be summarised in four main steps:

1. Sound design: the user authors a number of sounds by editing a set of salient synthesis parameters;
2. Agent exploration: the agent proposes a new mapping between the signals of the input device and the synthesis parameters based on previous feedback given by the user;
3. Play: the user plays with the synthesiser using the input device and the mapping proposed by the agent;
4. Human feedback: the user gives feedback to the agent.

In step 2, if no feedback was previously given the agent starts with a random mapping. Steps 3 and 4 are repeated until the user has found as many interesting motion-sound mappings as they like. The following subsections will describe the system architecture and a typical workflow.

It is worth noting that, differently from most IML applications for gestural interaction, there is not a gesture design step during which the performer records some sample sensor data for training the system. This is perhaps one of the most obvious differences between the IML and AIML paradigms. In an AIML workflow, the sample sensor data used for training the model is provided by the artificial agent, whereas the user gives feedback to the agent interactively while playing the resulting gesture-sound mappings.

Federico Ghelli Visi | mail@federicovisi.com | School of Music in Piteå, Luleå University of Technology, Sweden
Atau Tanaka | a.tanaka@gold.ac.uk | Goldsmiths, University of London, UK

### 6.2  AIML System Architecture

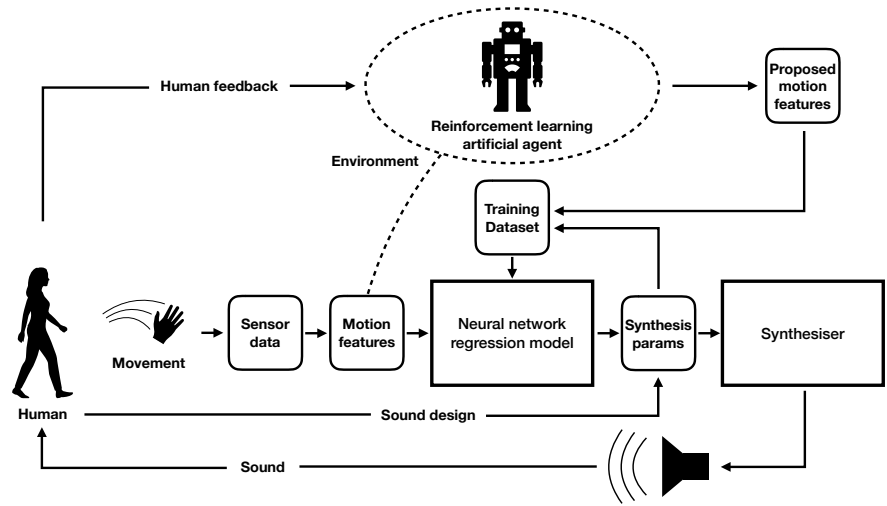

**Fig. 2.** Architecture of an Assisted Interactive Machine Learning system.

The architecture of the system is schematised in Figure 2. Motion features are stored in a vector and sent to a regression model created using a neural network. This was implemented in Max using the `rapidmax` object [66], an external built using RapidLib [67, 68], a set of software libraries for IML applications in the style of Wekinator [69]. These features also represent the dimensions of the environment in which the artificial agent operates. By exploring this feature space following the user's feedback, the agent proposes a set of motion features to be paired with the synthesis parameters defined by the user during the sound design step. This becomes the dataset used to train the neural network. The resulting regression model maps the incoming sensor data to sound synthesis parameters.

### 6.3  AIML Workflow

The four main steps of the interactive collaboration between the human performer and the artificial agent are schematised in Figure 3.

***1. Sound design*** In this first step, the user defines a number of sounds by manipulating a set of synthesis parameters. This process may differ depending on the synthesiser chosen and which synthesis parameters are exposed to the user in this step. In the first version of the system using the sample-based synthesiser described in section 5.1, the sounds are defined by manipulating six parameters

Federico Ghelli Visi | mail@federicovisi.com | School of Music in Piteå, Luleå University of Technology, Sweden
Atau Tanaka | a.tanaka@gold.ac.uk | Goldsmiths, University of London, UK

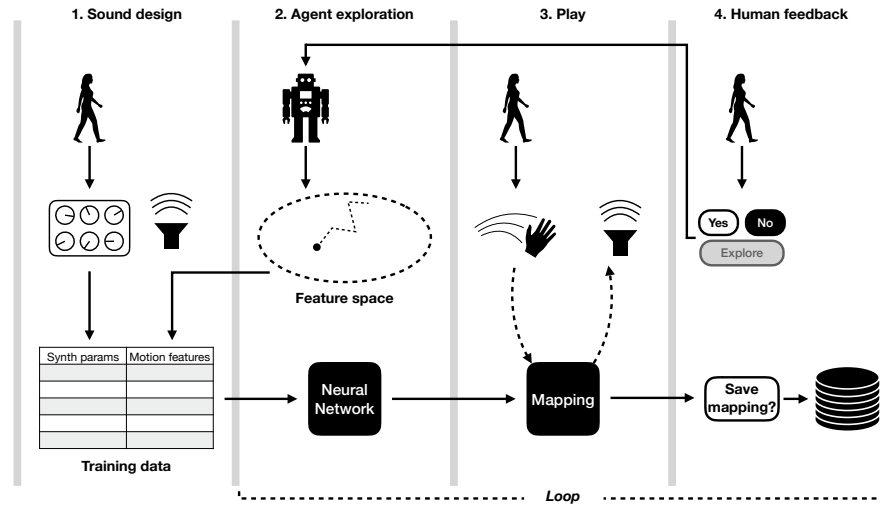

**Fig. 3.** The prototypical Assisted Interactive Machine Learning workflow.

(playback speed, pitch shift, start time, duration of the sample selection, filter cutoff frequency and resonance). Here, the user defines the parameters of four sounds that will be used to train a neural network in step 2 and perform regression in step 3. The sounds designed in the sound design step will thus act as timbral anchor points that define a space for interpolation and extrapolation of new sounds.

*2. Agent exploration* The dimensions of the environment explored by the agent are defined by the motion features extracted from the raw sensor data for each of the sound presets. Thus, at the end of the exploration step, the agent returns a vector with a set of input features for each of the sound synthesis parameters sets defined in the sound design step. This means that in the case of the version of the system using a 2D accelerometer, the agent will return four 2D vectors. These will be automatically paired with the synthesis parameters to train a neural network and create a regression model, which will be used in the following step to map live incoming sensor data to sound synthesis.

*3. Play* In this step, the user is free to play with and explore the resulting gesture-sound mapping for however long they like. Given that the regression models allow both interpolation and extrapolation of the input sound synthesis data, this step also allows to explore the timbral possibilities of the synthesiser while playing the mapping.

*4. Human feedback* After playing with the mapping, the user may give feedback to the artificial agent through a purposely designed interface. We adopted

Federico Ghelli Visi | mail@federicovisi.com | School of Music in Piteå, Luleå University of Technology, Sweden
Atau Tanaka | a.tanaka@gold.ac.uk | Goldsmiths, University of London, UK

the concepts of *guiding feedback* and *zone feedback* implemented in the agent designed by Scurto et al. [62]. Guiding feedback is a binary evaluation of the actions performed by the agent, or the direction of its exploration of the feature space. Zone feedback is instead an evaluation of the area of the feature space the agent is currently exploring. For example, a negative guiding feedback would change the direction of the agent's trajectory in the feature space, whereas a negative zone feedback would immediately transfer the agent to a different region of the space.

In our system, the user can give positive or negative guiding feedback to the agent about the proposed mapping. This feedback guides the direction of the next explorations of the feature space, and thus affects the next mappings proposed by the agent. In addition, the user can tell the agent to move to a different area of the feature space by means of a negative zone feedback. This will likely result in a new mapping that is considerably different from the previous one. In practice, this could be useful for trying something new once one is satisfied with the mappings proposed by the agent after a few guiding feedback iterations. In fact, whereas negative guiding feedback results in adjustments to the mappings currently being proposed by the agent, negative zone feedback triggers the exploration of a new area of the feature space, thereby exploring new mapping possibilities. Finally, users can save mappings, which can be retrieved later for performance or as material to be further refined using other interaction design approaches.

## 7   In Practice: IML Techniques in Musical Pieces

The sections that follow will describe how the techniques we outlined so far were employed by the authors for the development of their own musical pieces. The four pieces we selected showcase how these methods may be deployed to aid certain expressive intentions. Through their use in artistic practice, some of the creative affordances of IML paradigms become clearer, showing how certain creative processes may be facilitated and exposing strength and limitations of specific techniques.

### 7.1   Wais (Tanaka)

*Wais* (2019) is an homage to Michel Waisvisz, his work at the studio STEIM in Amsterdam and his performances on the instrument, *The Hands*. On one arm a short recording of a Waisvisz performance is articulated. On the other, an electronic music track, *Delull* by Tanaka. These two sources are granulated and placed in counterpoint. Two neural networks create independent regression models associating static posture and sound grain for each source. Once in "test" mode, these models take dynamic gesture, deconstructing the two prior works into a single improvisation.

Gesture is captured by one Myo sensor armband [70] on each forearm, providing 8 EMG channels, IMU quaternions, and combinatorial features resulting in 19 total muscle tension and movement features each from the left and right

Federico Ghelli Visi | mail@federicovisi.com | School of Music in Piteå, Luleå University of Technology, Sweden
Atau Tanaka | a.tanaka@gold.ac.uk | Goldsmiths, University of London, UK

arms. An instance of the synthesiser described in section 5.1 is associated with each sensor armband, allowing an independent sound buffer to be articulated by each arm.

The gesture input and sound synthesis output are associated by means of a neural network regression algorithm (see section 4.2), one for each arm. The performance consists of three sections, first to audition the unaltered source samples, second to train the neural network, and third to explore the trained model. At the beginning of the performance, the regression models are empty. The source sounds are played from the beginning, going up to 5 minutes for Tanaka's recording and 19 seconds looped of Waisvisz's recording. The overall summed RMS muscle tension for each arm initially modulates the amplitude of each recording, allowing the two musical voices to be articulated in a direct, gross manner. This section familiarises the listener with the original source materials.

In section two, a series of four granular regions and filter settings for each voice are associated with four static postures for each arm, with the gesture input recorded to establish a training set. This is done in performance as a sequence of events to set the synths to each pre-composed sound and prompt the performer to adopt a pose for each. The overall amplitude continues to be modulated by the summed muscle RMS. So, while each posture for the training set is static, the music continues to be articulated in a continuous manner through muscle tension. This creates a continuation of the first section where four segments of each voice are chosen as a way to zoom into segments of the original recordings. The eight segments are called up in an alternating for each arm through a rhythmic timing aided by a foot pedal push button to advancing and prompting the performer to each subsequent pose. After the training set of example poses and associated target sounds are recorded, the two neural networks are trained to produce a regression model.

The regression model is put into test mode for the third section of the piece and all three components of the work – 38 total dimensions of gesture, the two neural networks, and 12 total dimensions of sound synthesis output – come to life in dynamic interplay. The performer explores the gesture-sound space through continuous movement. He may approach or go through the poses from the training to see if the precomposed segments are recalled. He may explore multimodal decomposition and recombining of the source poses, perhaps striking a posture from one pose in space to recall IMU data for that pose, combined with muscle tension from another pose. This exploration is fluid, comprised of continuous gesture that dynamically goes in between and beyond the input points from the training set. The result is a lively exchange of the two musical voices, with the granular synthesis and filters constantly shifting in ways unlikely to be possible with manual parameter manipulation or direct mapping.

### 7.2   11 degrees of Dependence (Visi)

*11 Degrees of Dependence* (2016) is a composition for saxophone, electric guitar, wearable sensors, and live electronics that makes use of ML for continuously mapping the movements of the musicians to sound synthesis based on physical models and granular synthesis.

The piece explores the relationship between the performers and their instruments, focusing on the constraints that instrumental practice imposes on body movement and a topological interpretation of the musician's kinesphere [71]. The score includes symbols to notate movements, designed to be easily interpreted by musicians familiar with standard notation.

The piece is a duet for alto or soprano sax and electric guitar tuned in Drop C (open strings tuned CGCFAD from low to high). The sax player and the guitarist each wear two Myo armbands to control the physical model whereas the guitarist wears the same devices to control granular synthesis and an electroacoustic resonator placed on the guitar headstock. Parameter mapping is done using a supervised learning workflow based on SVMs. The data from the lateral (pitch) and longitudinal (roll) axes of the magnetometer are used as input to train the ML model. Four 'postures' are then defined for both musicians. In the case of the sax player, these are:

– a 'default' performance position (named 'Rest') with arms comfortably by the side of the chest,
– gently leaning back, raising the saxophone with the elbows slightly open (named 'Open'),
– leaning to the left with the right elbow slightly pointing outwards (named 'Left'),
– leaning to the right with the left elbow slightly pointing outwards (named 'Right').

During the training phase, each posture is coupled with a set of synthesis parameters of the flute physical model. The Rest posture is paired with a clean sound with a clear fundamental frequency, the Open posture with a louder sound rich of breath noise, the Left posture adds overtones, and the right posture with a flutter tongued 'frullato' sound. In performance (testing phase, in ML terms), the synthesis parameters are continuously interpolated using the output likelihoods of the classifier as interpolation factors. This synthesised wind instrument sounds are designed to blend with the saxophone sound to generate a timbre with both familiar and uncanny qualities. The pitch played by the flute model is a C1, which is also the tonic of the piece. The amount of noise fed into the physical model (or breath pressure) is controlled by the sum of the EMG MAV values of both arms. This implies that the amount of synthesised sound is constrained by the movement of the fingers operating the saxophone keys. Notes that require more tone holes to be closed – such as low notes for example – cause more muscular activity and thus louder sounds from the physical model. This design choice adds a component of interdependent, semi-conscious control to the performance creating a tighter coupling between the sounds of the saxophone and those of the flute model.

*11 Degrees of Dependence* is structured in 3 parts, each of which contains scored themes at the beginning and the end a middle improvised section. The full score of the alto saxophone part can be found in appendix of [72]. The score

Federico Ghelli Visi | mail@federicovisi.com | School of Music in Piteå, Luleå University of Technology, Sweden
Atau Tanaka | a.tanaka@gold.ac.uk | Goldsmiths, University of London, UK

adopts conventional notation along with some custom symbols (printed in red) used to notate movement. While the symbols indicate at which point in time the posture should be reached, the red lines show how the transition between the different postures should be articulated. These lines resemble other lines commonly found in conventional music notation. A straight line between two symbols means that the performer should start from the posture represented by the first symbol and progressively move towards the posture represented by the second symbol. The movement resulting from the transition between the postures should end in correspondence with the second symbol, thus following the rhythmic subdivision indicated in the staff. This is similar to a glissando, also notated using straight lines between note heads. A curved line between the posture symbols works instead analogously to a legato, meaning that the indicated posture quickly tied with the following one. The score is where affordances and constraints of the agencies involved in the piece coalesce: each posture is represented by a symbol and corresponds to a class of the ML classifier, body movements occur in-between postures, causing sound synthesis to move in-between predefined parameter sets. At the centre of these interdependent agencies we find the bodies of the musicians, and their embodied relationships with their instruments.

### 7.3 Delearning (Tanaka)

*Delearning* (2019) takes as its source a work by Tanaka for chamber orchestra, DSCP, as sound corpus for analysis and subsequent neural network regression. Feature extraction of arm poses are associated with audio metadata and used to train an artificial neural network. The algorithm is then put into performance mode allowing the performer to navigate a multidimensional timbre space with musical gesture.

This piece puts in practice the technique we describe in Zbyszyński et al. [21], where multimodal EMG and IMU sensing is used in conjunction with corpus based concatenative sound synthesis (CBCS) to map 19 dimensions of incoming gesture features by a regression model to 19 dimensions of audio descriptors.

The nineteen gesture features are taken from the right forearm and are: IMU quaternions (4 dimensions), angular velocity (4 dimensions), 8 channels of Bayes filtered EMG, total summed EMG (1 dimension), and a separation of all EMG channels on the perimeter of the forearm to horizontal and vertical tension (2 dimensions).

The nineteen target audio descriptors are grain duration followed by the means and standard deviations of frequency; energy; periodicity; autocorrelation coefficient; loudness; spectral centroid, spread, skew and kurtosis.

The gesture input feature space is mapped to the target audio feature space by means of a neural network that creates a regression model associating gesture as performed and sound synthesis output.

The source audio is a recording of an 18 minute piece for chamber orchestra of mixed forces. The work was chosen as it contains a diverse range of timbres and dynamics all while being musically coherent. Before the performance, the recording is analysed to generate the audio descriptors. The recording is read from beginning to end and is segmented by transient onset detection into grains. This generates 21,000 grains over the course of the duration of the recording, making the average grain 50ms in duration.

The composition consists of five points in the original piece that have been selected to be associated with performance postures. The EMG and IMU sensors on the right arm feed the neural network, while EMG amplitude from the left arm modulated overall synthesiser amplitude and IMU quaternions modulate, at different points in the composition, filtering and spatialisation.

The performance begins with the analysed recording, but with an empty regression model. The first grain is heard, and the performer adopts a posture to associate with it, and records that as training data into the neural network. This continues for the two subsequent grains, at which point the training set consists of gesture features of the three poses associated with the audio descriptors of the three grains. The neural network is trained, and it then put into test mode and the performer explores this gesture-timbre space through fluid, dynamic gesture. The regression model takes gesture feature input to report a set of audio descriptors to the synthesiser. The synthesiser applies a K-nearest neighbour algorithm to find a grain in the corpus that has the closest Euclidean distance to the look up features.

In the next section of the piece, the neural network is put back into training mode, and a fourth grain is introduced and associated with a fourth posture. This pose is recorded as an extension to the existing training set, putting in practice the IML paradigm of providing more examples. The neural network is retrained on this enhanced data set and put in performance mode for further free exploration by the performer.

This is repeated with the fifth and final grain to extend the regression model one last time to model the data representing 5 poses associated with five audio grains. This creates a musical structure where the gesture-timbre space becomes richer and more densely populated through the development of the piece.

### 7.4 "You Have a New Memory" (Visi)

*"You Have a New Memory"* (2020) [73] makes use of the AIML interaction paradigm (see section 6) to navigate a vast corpus of audio material harvested from the messaging applications, videos, and audio journals recorded on the author's mobile phone. This corpus of sonic memories is then organised using audio descriptors and navigated with the aid of an artificial agent and RL. Feedback to the agent is given through a remote control, while embodied interaction with the corpus is enabled by a Myo armband.

Sonic interaction is implemented using CBCS (see section 5.2). The approach is further refined by adopting the method based on self-organising maps proposed by Margraf [74], which helps handling the sparseness of heterogenous audio corpora.

In performance, the assisted exploration of sonic memories involves a embodied exploration of the corpus the entails both a search of musical motives and

Federico Ghelli Visi | mail@federicovisi.com | School of Music in Piteå, Luleå University of Technology, Sweden
Atau Tanaka | a.tanaka@gold.ac.uk | Goldsmiths, University of London, UK

timbres to develop gestural musical phrases, as well as the intimate, personal exploration of the performer's recent past through fragments of sonic memories emerging from the corpus following the interaction with the agent. The juxtaposition of sounds that are associated with memories from different periods may guide the performer towards an unexpected introspective listening that co-inhabits the performance together with a more abstract, sonic object-oriented reduced listening [14]. The shifting between these modalities of listening influences the feedback given to the agent, which in return alters the way the performer interacts with the sonic memories stored in the corpus.

The title of the piece – *"You Have a New Memory"* – refers to the notifications that a popular photo library application occasionally send to mobile devices to prompt their users to check an algorithmically generated photo gallery that collects images and videos related to a particular event or series of events in their lives. These collections are ostensibly compiled by algorithms that extract and analyse image features, metadata (e.g. geotags), and attempt to identify the people portrayed on the photos [75].

The piece aims at dealing with the feelings of anxiety associated with the awareness that fragments of one's life are constantly turned (consciously or not) into data that is analysed and processed by unattended algorithms, whose inner workings and purposes are often opaque. The piece then is also an attempt at actively employing similar algorithms as a means of introspection and exploration. Rather than passively receiving the output of ML algorithms dictating when and how one's memories are forming, here the algorithms are used actively, as an empowering tool for exploring the complexity outlined by the overwhelming amount of data about ourselves that we constantly produce.

## 8 Conclusion

The purpose of this chapter was to provide an overview of the solutions, challenges, needs, and implications of employing IML techniques for analysing and designing musical gestures. The research field is still rapidly developing, and the topics we touched in the previous sections may give an idea of the interdisciplinary effort required for advancing research further. Advances in the field require an interdisciplinary perspective as well as a methodology encompassing basic research enquiry, the development of tools, their deployment in artistic practice and an analysis of the impact such techniques have on one's creative process. Learning more about the use of ML in music has manifold implications, stretching way beyond the music domain. As ML technologies are used to manage more and more aspects of everyday life, working along the fuzzy edges of artistic practice – where tasks are often not defined in univocal terms and problems are, and need to be left, open to creative solutions – becomes a laboratory in which we understand how to claim and negotiate human agency over data systems and algorithms. Understanding ML as a tool for navigating complexity that can aid musicians' creative practice may contribute to the advancement of these techniques as well as to their demystification and broader adoption by artists, researchers as well as educators, thereby becoming sources of empowerment and inspiration.

## 9 Acknowledgements

The *GEMM))) Gesture Embodiment and Machines in Music* research cluster is funded by Luleå University of Technology, Sweden.

Some of the research carried out at Goldsmiths, University of London described in this chapter has received funding from the European Research Council (ERC) under the European Union's Horizon 2020 research and innovation programme (Grant agreement No. 789825 – Project name: BioMusic).

The authors would like to acknowledge Micheal Zbyszyński and Balandino Di Donato for their contributions to the BioMusic project and the discussions and sharing of ideas that led to some of the work presented in this chapter.

Federico Ghelli Visi | mail@federicovisi.com | School of Music in Piteå, Luleå University of Technology, Sweden
Atau Tanaka | a.tanaka@gold.ac.uk | Goldsmiths, University of London, UK

Federico Ghelli Visi | mail@federicovisi.com | School of Music in Piteå, Luleå University of Technology, Sweden
Atau Tanaka | a.tanaka@gold.ac.uk | Goldsmiths, University of London, UK
Author's accepted manuscript, to appear as a chapter in "Handbook of Artificial Intelligence for Music: Foundations, Advanced Approaches, and Developments for Creativity", edited by E. R. Miranda. Cham: Springer Nature, 2021.
26      Visi & Tanaka30. D. Glowinski, N. Dael, A. Camurri, G. Volpe, M. Mortillaro, and K. Scherer, "Toward a Minimal Representation of Affective Gestures," *IEEE Transactions on Affective Computing* **2**, pp. 106–118, apr 2011.
31. K. Hachimura, K. Takashina, and M. Yoshimura, "Analysis and evaluation of dancing movement based on LMA," in *ROMAN 2005. IEEE International Workshop on Robot and Human Interactive Communication, 2005.*, **2005**, pp. 294–299, IEEE, 2005.
32. F. Visi, R. Schramm, and E. Miranda, "Gesture in performance with traditional musical instruments and electronics," in *Proceedings of the 2014 International Workshop on Movement and Computing - MOCO '14*, pp. 100–105, ACM Press, (New York, NY, USA), 2014.
33. M. Müller, *Information retrieval for music and motion*, Springer, Berlin, Germany, 2007.
34. B. Burger and P. Toiviainen, "MoCap Toolbox – A Matlab toolbox for computational analysis of movement data," in *Proceedings of the 10th Sound and Music Computing Conference*, R. Bresin, ed., pp. 172–178, KTH Royal Institute of Technology, (Stockholm, Sweden), 2013.
35. R. Schramm and F. G. Visi, "Periodic quantity of motion repository." https://github.com/schramm/pqom/, 2019. Accessed: 2020-06-16.
36. T. D. Sanger, "Bayesian Filtering of Myoelectric Signals," *Journal of Neurophysiology* **97**, pp. 1839–1845, feb 2007.
37. A. Phinyomark, P. Phukpattaranont, and C. Limsakul, "Feature reduction and selection for EMG signal classification," *Expert Systems with Applications* **39**, pp. 7420–7431, jun 2012.
38. Z. Arief, I. A. Sulistijono, and R. A. Ardiansyah, "Comparison of five time series EMG features extractions using Myo Armband," in *2015 International Electronics Symposium (IES)*, pp. 11–14, IEEE, sep 2015.
39. K. S. Kim, H. H. Choi, C. S. Moon, and C. W. Mun, "Comparison of k-nearest neighbor, quadratic discriminant and linear discriminant analysis in classification of electromyogram signals based on the wrist-motion directions," *Current Applied Physics* **11**, pp. 740–745, may 2011.
40. J. Kaiser, "On a simple algorithm to calculate the 'energy' of a signal," in *International Conference on Acoustics, Speech, and Signal Processing*, **1**, pp. 381–384, IEEE, 1990.
41. E. Kvedalen, *Signal processing using the Teager Energy Operator and other nonlinear operators*. Master thesis, Universitetet i Oslo, may 2003.
42. S. Solnik, P. DeVita, P. Rider, B. Long, and T. Hortobágyi, "Teager-Kaiser Operator improves the accuracy of EMG onset detection independent of signal-to-noise ratio." *Acta of bioengineering and biomechanics* **10**(2), pp. 65–8, 2008.
43. R. Bachu, S. Kopparthi, B. Adapa, and B. Barkana, "Voiced/Unvoiced Decision for Speech Signals Based on Zero-Crossing Rate and Energy," in *Advanced Techniques in Computing Sciences and Software Engineering*, pp. 279–282, Springer Netherlands, Dordrecht, 2010.
44. B. Caramiaux, M. Donnarumma, and A. Tanaka, "Understanding Gesture Expressivity through Muscle Sensing," *ACM Transactions on Computer-Human Interaction* **21**, pp. 1–26, jan 2015.
45. M. Mehryar, R. Afshin, and T. Ameet, *Foundation of Machine Learning, Second Edition*, MIT Press, 2018.
46. F. Visi, B. Caramiaux, M. Mcloughlin, and E. Miranda, "A Knowledge-based, Data-driven Method for Action-sound Mapping," in *NIME'17 – International Conference on New Interfaces for Musical Expression*, 2017.

Federico Ghelli Visi | mail@federicovisi.com | School of Music in Piteå, Luleå University of Technology, Sweden
Atau Tanaka | a.tanaka@gold.ac.uk | Goldsmiths, University of London, UK

Federico Ghelli Visi | mail@federicovisi.com | School of Music in Piteå, Luleå University of Technology, Sweden
Atau Tanaka | a.tanaka@gold.ac.uk | Goldsmiths, University of London, UK